\pdfoutput=1

\documentclass[11pt]{article}

\usepackage[final]{acl}

\usepackage{times}
\usepackage{latexsym}
\usepackage{amsmath}
\usepackage{amssymb}
\usepackage{float}
\usepackage{balance} 
\usepackage{afterpage} 

\AtBeginEnvironment{verbatim}{\small} 
\usepackage[T1]{fontenc}

\usepackage[utf8]{inputenc}

\usepackage{microtype}

\usepackage{inconsolata}

\usepackage{graphicx}

\title{Linear Relational Decoding of Morphology in Language Models}

\author{Eric Xia\\
	    Brown University\\
	    {\tt eric\_xia@brown.edu}
	  \And
	Jugal Kalita\\
  	University of Colorado Colorado Springs\\
  {\tt jkalita@uccs.edu}}

\begin{document}
\maketitle
\begin{abstract}
A two-part affine approximation has been found to be a good approximation for transformer computations over certain subject-object relations. Adapting the Bigger Analogy Test Set, we show that the linear transformation $W\textbf{s}$, where $\textbf{s}$ is a middle layer representation of a subject token and $W$ is derived from model derivatives, is also able to accurately reproduce final object states for many relations. This linear technique is able to achieve 90\% faithfulness on morphological relations, and we show similar findings multi-lingually and across models. Our findings indicate that some conceptual relationships in language models, such as morphology, are readily interpretable from latent space, and are sparsely encoded by cross-layer linear transformations.
\end{abstract}

\section{Introduction}
\begin{figure*}[htbp]
    \centering
    \includegraphics[width=1\linewidth]{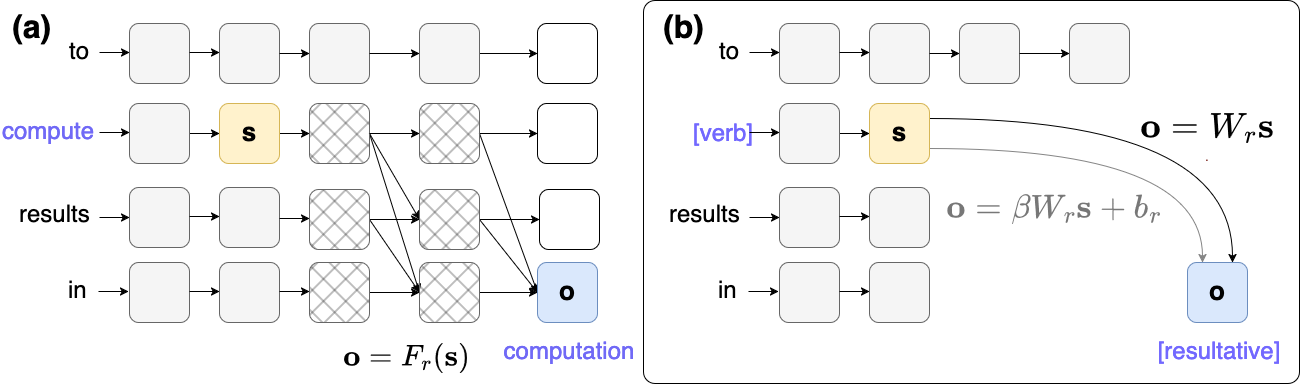}
    \caption{As seen in \textbf{(a)}, transformers resolve subject-object relations in a highly nonlinear fashion. As seen in \textbf{(b)}, both affine and linear approximators of the subject-object map $F_r(\textbf{s})$ are demonstrated to be highly effective over relations such as morphology.}
    \label{fig:enter-label}
\end{figure*}
Large language models display impressive capabilities for factual recall, which commonly involve relations between entities (\citealt{Brown:20}). Recent work has shown that affine transformations on subject representations can faithfully approximate model outputs for certain subject-object relations (\citealt{Her:23}). Identifying transformer approximators is an important area of study, with applications in model training and editing.

The contributions of this paper are twofold. We reproduce and extend existing research. Specifically, we apply affine Linear Relational Embedding (LRE) method to novel diverse relational categories, including derivational and inflectional morphology, encyclopedic knowledge, and lexical semantics. By doing so, we confirm the efficacy of the affine technique. We show that relational approximation can be applied to adapted analogical datasets, and demonstrate relational approximation for a broad range of linguistic phenomena.

At the same time, this work makes a key contribution to research on relational representation in model latents. We show that for different relations, additive and multiplicative mechanisms play complementary roles in affine approximation. We find that an analogue to the original linear relational embedding developed by \cite{Pac:01}, using a single multiplicative operator, is effective within specific relations. In particular, linear approximation within contexts relating morphological forms reaches near-equivalent level of faithfulness to the affine LRE. We test faithfulness over eight different languages and find that this equivalence holds cross-typologically.

\section{Related Work}
Much work in machine learning has focused on learning concept representations with hierarchical structure. Relations between representations in concept spaces have been modeled successfully by both linear multiplicative and additive operations.  \\ \\
\textbf{Multiplicative.} \citet{Pac:01} introduced the concept of the linear relational embedding for learning relational knowledge from triples ($a$, $R$, $b$). Concepts such as $a$ and $b$ are represented as $n$-length vectors, while relations such as $R$ are represented as $n \times n$ matrices, akin to distributional models of compositional semantics proposed by \citet{Coecke:10}. 
\\ \\
\textbf{Additive.} \cite{Mik:13} used linear operations in word vector space derived from context-predictive neural nets, demonstrating a correspondence between directional binary relations (e.g. male-female, country-capital, verb tense) and the addition of embedding vectors. Later work found inflectional relations were better captured than derivational ones, and encyclopedic relations better than lexicographic ones. (\citealt{Glad:16,Vyl:16}). 

 
\section{Background}
\subsection{Transformer Computation}
In auto-regressive transformer language models, input text is converted to a sequence of tokens $t_1 \ldots t_n$, which are subsequently embedded as $x_1 \ldots x_n \in \mathbb{R}^d$ by an embedding matrix. They are then passed through $L$ transformer layers, each composed of a self-attention layer and an multi-layer perceptron (MLP) layer. In GPT-J, the representation $x^l_i$ of the $i^\text{th}$ token at layer $l$ is obtained as:
$$x^l_i = x^{l-1}_i + a^{l}_i + m^{l}_i$$
where $\mathrm{a}^l_i$ is multi-headed Key-Value Query attention over $x^{l-1}$(\citealt{Vas:17}) and $m^l_i$ is the $i^\mathrm{th}$ output of the $l ^\mathrm{th}$ MLP sublayer. In this case, the output of the $l$-th MLP sublayer for the $i$-th representation depends on $x^{l-1}_i$, rather than $a^{l}_i + x^{l-1}_i$ (\citealt{GPT-J}). The final prediction $t_{n+1}$ is then determined by the final hidden state $x_n$ passed through a decoder head $D$, which consists of a linear layer and softmax to a token vocabulary: $t_{n+1} = \underset{t}{\mathrm{argmax}} \ D(x^L_n)_t $.
\subsection{Relational Representation}
Throughout this paper, we will focus on the subject-object relationship as expressed through a single fixed context. Following prior work (\citealt{Meng:22b,Geva:23}) that the last token state of a subject in middle layers are strongly casual on predictions (e.g. "Needle" in "The Space Needle"), we are interested in utilizing the gradient between the last token position of the subject $s$ at an intermediate layer, and the object prediction state $o$. 

\section{Approach} 
\subsection{Problem Statement}
We first consider what it means for a context to express a relation. Many statements can be expressed in terms of a subject, relation, and object (\textit{s,r,o}). For instance, the statement \textit{Miles Davis plays the trumpet} expresses a relation $F_r$, connecting the subject $s$ (\textit{Miles Davis}) to the object $o$ (\textit{trumpet}): $F_r(s) = o$. We can then relate new subjects to objects: $F_r(\textit{Jimi Hendrix}) = \textit{guitar}$ and $F_r(\textit{Elton John}) = \textit{piano}$. $F_r$ is an inductive mechanism, from which statements relating subject and object pairs can be obtained. We are interested in how a language model implements this abstraction.
\noindent \\
\textbf{Affine LRE.} As a starting point, we look at the affine linear relational embedding (LRE) method developed by \cite{Her:23}. The authors are able to approximate the transformer's relational function $F_r(s)$ with the affine approximator $\textrm{LRE}(s)$, such that when applied to novel subjects, they reproduce LM object predictions. 

\begin{figure*}[htbp]
    \centering
    \includegraphics[width=1\linewidth]{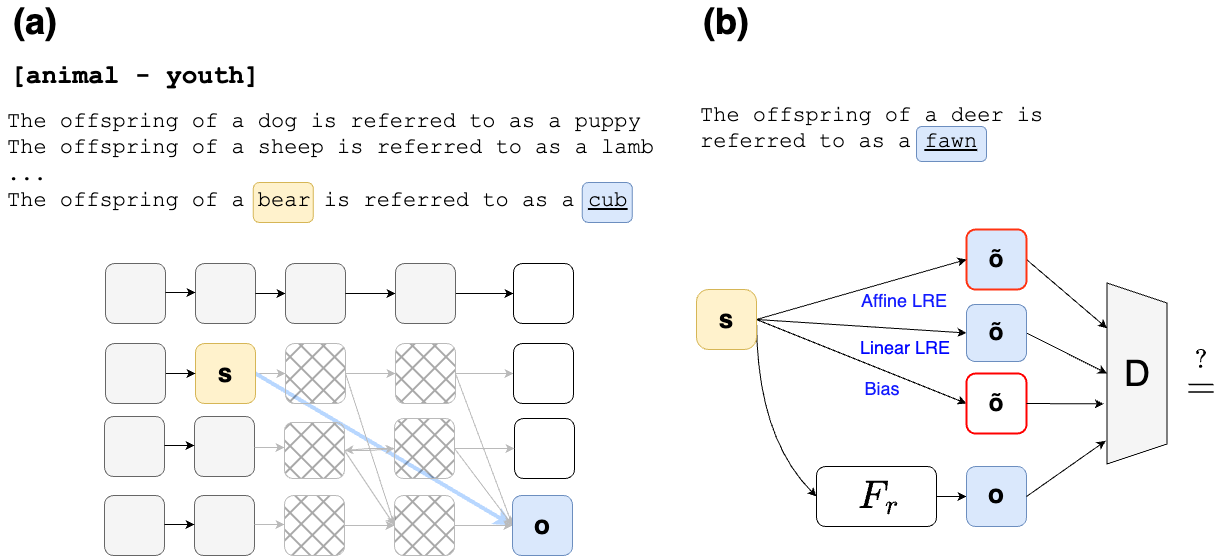}
    \caption{In \textbf{(a)}, we first assemble approximators from trained model Jacobians between middle-layer subject states and the final-layer object state. Then, in \textbf{(b)}, we evaluate approximated tokens against transformer computations.
}
    \label{fig:methodology}
\end{figure*}

The object retrieval function from a subject with a fixed relational context, $o = F_r(s)$, is modeled to be a first-order Taylor approximation of $F_r$ about a number of subjects $s_1 \ldots s_n$. For $i = 1 \ldots n$:
 \begin{align*}
F_r(s) &\approx F_r(s_i) + W_r(s - s_i) \nonumber \\
       &= F(s_i) + W_rs - W_rs_i \nonumber \\
       &= W_rs + b_r,  \nonumber \\
\text{where } b_r &= F_r(s_i) - W_rs_i \nonumber
\end{align*}

In a relational context, a model may rely heavily on a singular subject state to produce the object state. Accordingly, the Jacobian matrix of derivatives between vector representations of the subject and object is hypothesized to serve as $W_r$. For a fixed relation, they calculate the mean Jacobian and bias between $n$ enriched subject states $\textbf{s}_1 \ldots \textbf{s}_n$ and outputs $F_r(\textbf{s}_1) \ldots F_r(\textbf{s}_n)$:

\begin{align*}
W_r &= \mathbb{E}_{\textbf{s}_i}\left[\left.\frac{\partial F_r}{\partial \textbf{s}}\right|_{\textbf{s}_i} \right] && \text{($d \times d$ matrix)} \\
b_r &= \mathbb{E}_{\textbf{s}_i}\left[\left. F_r(\textbf{s}) - \frac{\partial F_r}{\partial \textbf{s}} \; \textbf{s} \right|_{\textbf{s}_i} \right] && \text{($d$ vector)}
\end{align*}
\noindent

This yields a relational approximator capable of transforming a $j^\text{th}$ layer subject state $x^j_s = \textbf{s}$ \footnote{Following \cite{Meng:22}, both this paper and the affine LRE focus primarily on middle-layer states.} into the final object hidden state $x^L_o = \textbf{o}$  \footnote{Note the introduction of a $\beta$ scaling parameter. The authors claim the affine LRE is limited by layer normalization: the \textbf{s} representation is normalized before contributing to \textbf{o}, and \textbf{o} is normalized before token prediction by the LM head, resulting in a mismatch in the scale of the output approximation. We find that this conclusion is supported by empirical evidence from linear projections.}:
 $$\textbf{o} \approx \textrm{LRE(\textbf{s})} = \beta W_r \textbf{s} + b_r$$
For instance, $\textbf{s}$ may be the hidden state of the 7$^\text{th}$ layer at the subject token, and $\textbf{o}$ the hidden state of the 26$^\text{th}$ (last) layer at the object token, e.g. the next-token prediction state.

\begin{figure*}[h!]
    \centering
        \includegraphics[width=1\linewidth]{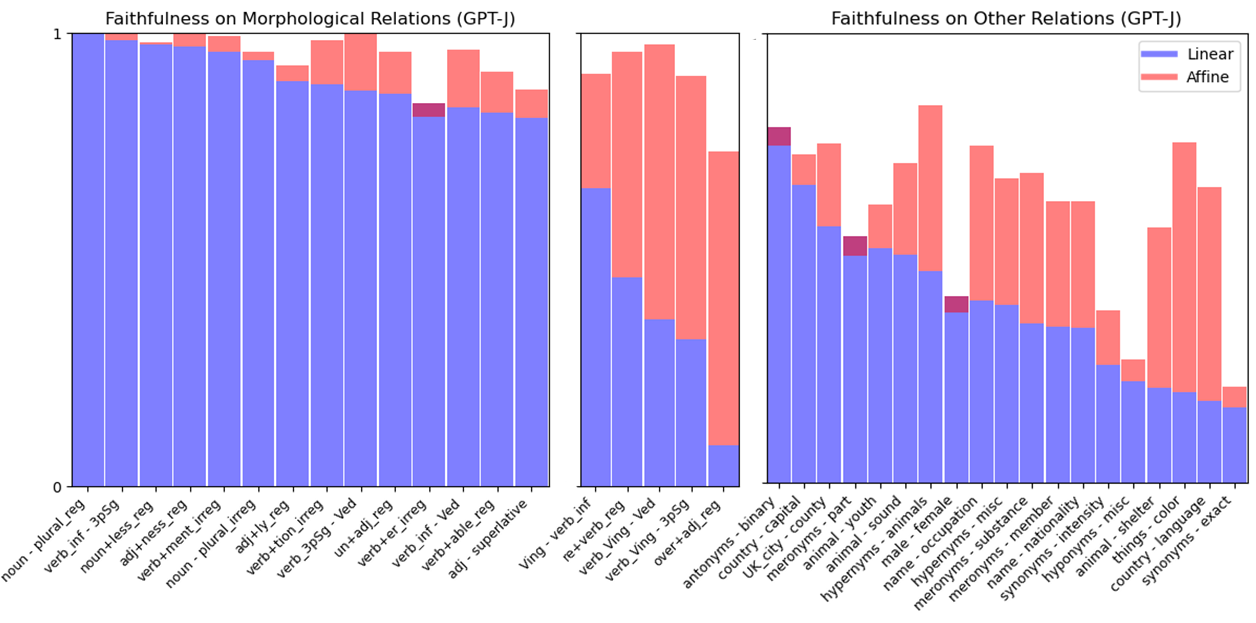}
    \caption{Comparing affine \& linear LREs on GPT-J reveals many morphological relations are linearly approximable. With the exception of prefix and active form subjects, semantic and encyclopedic relations benefit more from the affine LRE than morphology. For subject layers 3-9, the best performing approximation is averaged ($n = 4$).}
    \label{fig:faithfulness-comparison}
\end{figure*}
\noindent
\textbf{True Linear Encoding. }The affine LRE diverges from the linear relational embedding introduced by \cite{Hinton:86}, in introducing a bias $b_r$ and scaling term $\beta$. While linearity is assumed in \citet{Her:23} by calculating $W_r$ and $b_r$ from $\mathbb{E}_{s_i}$ over $i = 1 \ldots n$, using a Taylor approximation makes a weaker assumption, simply that the subject-object relation $F_r$ is differentiable. With linearity, we would expect the following:
\begin{align*}
 \textbf{o} &\approx F_r'(s_i)\textbf{s} \\
            &= W_r \textbf{s}
\end{align*}
In this case, the linear approximation over $\textbf{s}_1 \ldots \textbf{s}_n$ within the same relation would be the mean Jacobian. If this approximation generalizes to unseen objects, it would indicate the presence of a linear subject-object map.


\subsection{Introducing New Relations}
Analogy is traditionally seen as a special case of role-based relational reasoning (\citealt{Stern:79}, \citealt{Gentner:83}, \citealt{Holy:12}), motivating the adaptation of analogical pairs to a relational setting. We choose to adapt the Bigger Analogy Test Set (BATS), originally introduced to explore linguistic regularities in word embeddings by \cite{Glad:16}. The dataset comprises forty different categories, each with fifty pairs of words sharing a common relation. The categories span inflectional morphology, derivational morphology, encyclopedic knowledge, and lexical semantics.

\subsection{Utilizing ICL}

As seen in Figure \ref{fig:methodology}, we adapt the relational pairs in BATS by introducing prompts which are compatible with each instance of the analogy. 

Following the procedure outlined in Hernandez \shortcite{Her:23}, we employ 8 in-context learning (ICL) examples for 8 different subject-object prompts for each relation. This allows us to obtain a Jacobian from the model computation which is most likely to exhibit the desired linear encoding.

We omit the subject-object pairs used in construction from the testing pool. We further restrict evaluation to the pairs for which the LM computation is successful in reproducing the object. \footnote{For both GPT-J and Llama-7b, nearly all examples fit this criteria.}

\begin{figure*}[htbp]
    \centering
        \includegraphics[width=1\linewidth]{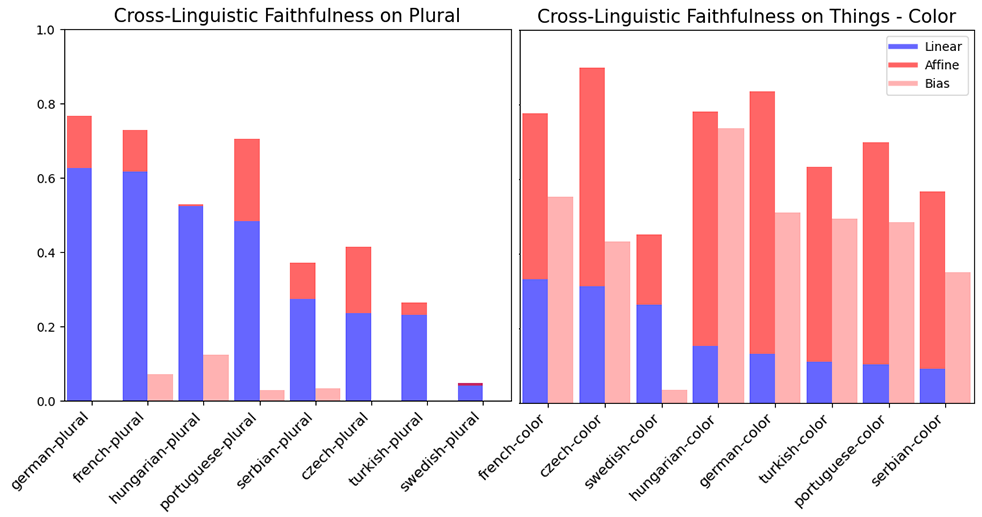}
    \caption{Evaluating languages present in Llama-7b reveal cross-typological linear encoding of morphology. Linear and affine LREs respectively score 56\% and 68\% on \textbf{[plural]} across German, French, Hungarian, and Portuguese. In contrast, on \textbf{[things - color]} relation the linear and affine techniques respectively score 19\% and 70\%. The Bias approximator scores 45\%, suggesting the affine approximation for \textbf{[things - color]} is primarily additive.}
    \label{fig:crossling-faith}
\end{figure*}

\subsection{Evaluating Operators}
After passing through the activation function in the decoder, the approximated object tokens should faithfully replicate the true LM output. \\ \\
\textbf{Affine LRE.} The original affine LRE is a two-step approximation involving both a weight term $W_r$ and bias term $b_r$, which are applied to the subject state $\textbf{s}$ to produce an approximated output state: $\tilde{\textrm{\textbf{o}}} = \text{LRE(\textbf{s})} = \beta W_r\textbf{s} + b_r 
    \label{LRE}$
\\ \\
\textbf{Linear LRE.} Our variants isolate the components of the LRE in order to inspect their contribution to the approximation. First, we define the linear LRE, a multiplicative operation. This is the subject hidden state \textbf{s} multiplied by the mean Jacobian for \textit{other subject-object pairs} to derive a final object state:
    $\tilde{\textrm{\textbf{o}}} = \text{Linear(\textbf{s})} = W_r\textbf{s}
    \label{Linear}$
\\ \\
\textbf{Bias.} Second, we define the Bias approximator, an additive operation. This approximator calculates $\tilde{\textrm{\textbf{o}}}$ by adding $b_r$, the mean difference between $W_r \textbf{s}$ and \textbf{o} for \textit{other subject-object pairs}, to \textbf{s}: $\tilde{\textrm{\textbf{o}}} = \text{Bias({\textbf{s})}} = \textbf{s} + b_r
\label{bias}$
\\ \\
Following \cite{Her:23}, we define \textit{faithfulness} of an approximator by the top-one token match rate. For token $t$ and decoder head $D$, we say an approximator is faithful if the top token approximation matches that of the LM:
$\label{eq:faithfulness} \underset{t}{\mathrm{argmax}} \ D(\textbf{o})_t \stackrel{?}{=} \underset{t}{\mathrm{argmax}} \ D(\tilde{\textbf{o}})_t$

\section{Results}

\subsection{The Linear LRE Faithfully Approximates Relations across Morphology}
We first evaluate relational approximators for the GPT-J model (\citealt{GPT-J}). We build approximators for likely subject hidden states (layers 3-9) and the final object state (layer 27) through the process outlined above. We then evaluate the approximators four times for each relation, and average the best cross-layer approximation.\footnote{There were two relations which were not tested on, \textbf{[adj+comparative]} and \textbf{[antonyms-gradable]}. This was due to preprocessing issues.} \footnote{For the LRE, we use $\beta=7$, which was found to be optimal for BATS. }

As seen in Figure \ref{fig:faithfulness-comparison}, the linear LRE achieves 90\% faithfulness across 14 morphology relations, while the affine LRE achieves a faithfulness of 95\%. In contrast, the linear LRE achieves 40\% faithfulness over non-morphological relations, while the affine LRE achieves 61\% faithfulness. This confirms the efficacy of the affine LRE found by \cite{Her:23}, while suggesting that some relations, e.g. morphology, may be encoded as truly linear.

 To show that the Jacobian is not only sufficient but also necessary, in Appendix Figure \ref{fig:bias} and Appendix Figure \ref{fig:translation} we compare the LREs against two additive approximations, Bias and {\scshape TRANSLATION}. TRANSLATION adds the mean difference between the subject and object states to each subject state. In both cases, we find that an additive operator is unable to reproduce morphology.


\subsection{Llama-7b Results}
GPT-J utilizes parallel MLP and attention layers, unlike many other language models. Consequently, it is possible the observed linearity does not generalize to different architectures. We repeat the procedure for Llama-7b, which like most LLMs utilizes sequential attention and feedforward layers (\citealt{LLAMA2}). In the Appendix Figure \ref{fig:faithfulness-llama}, we display similar results to Figure \ref{fig:faithfulness-comparison}; suggesting similar encoding mechanisms exist across models. 

\subsection{Cross-Linguistic Evidence}
We have shown that morphological relations in English are largely linearly decodable. However, these results may be limited to fusional-analytic languages with fewer unique affixes. For Llama-7b, we test Czech, French, German, Hungarian, Portuguese, Serbian, Swedish, and Turkish, each comprising significant portions of the training dataset. Hungarian and Turkish are both highly agglutinative. We create templates for one morphological (\textbf{[plural]}) and non-morphological relation (\textbf{[things - color]}). We evaluate approximators as above.

As seen in Figure \ref{fig:crossling-faith}, affine and linear approximators achieve similar results on \textbf{[plural]}, while the additive operation performs well on \textbf{[things - color]}. These results indicate a multiplicative linear relational embedding for certain morphological relations, independent of linguistic typology. The high performance of the additive Bias operator on \textbf{[things - color]} provides evidence for complementary additive and multiplicative mechanisms.

\section{Conclusion}
In this work, we have adapted a large relational dataset for testing transformer approximation. We formulate the transformer version of the linear relational embedding found in \cite{Pac:01} more precisely to be equivalent to a matrix-vector multiplication with the mean Jacobian. Surprisingly, we find this linear operation is able to model certain relations such as morphology nearly as well as the affine LRE. This suggests that certain conceptual relations surface linearly in the residual space of language models, and are sparsely encoded multiplicatively as opposed to additively. 

\section{Limitations}
Our experiments were conducted exclusively on GPT-J and Llama-7b due to hardware constraints, which limited the scope of our evaluations. However, smaller models serve as a likely proxy for studying the interpretability of transformer-based language models due to identical architectures.

Throughout the work, we assume linear transformations observed are employed in token prediction through the same mechanism as in explicit relational contexts. Existing literature in activation patching and editing indicates that subject enrichment occurs independently from surrounding contexts (\citealt{geva:21}), indicating that the relational embedding outlined here is consistent.

Unlike previous investigations of linear approximation, we did not investigate whether the faithfulness of the Jacobian approximation is associated with causality. Based on prior work which finds a consistent relationship between these variables (\citealt{Her:23}), these two measures appear correlated.

\subsubsection*{Acknowledgments}
The research done here was supported by the National Science Foundation under award number \#2349452. Any opinion, finding, or conclusion in this study is that of the authors and does not necessarily reflect the views of the National Science Foundation.

\bibliography{custom}

\clearpage
\appendix

\section{Reproducibility Statement}
The approximation code is based on the LRE repository (\citealt{Her:23}), and loads GPT-J and Llama-7b in half-precision. The code and dataset are available at \href{https://github.com/rkique/linear-morphology}{https://github.com/rkique/linear-morphology}. Experiments were run remotely on a workstation with 24GB NVIDIA RTX 3090 GPUs using HuggingFace Transformers.

\section{Evidence of non-stemmed forms}
As seen in Table \ref{tab:full-forms}, the linear LRE successfully replicates full forms for many derived object states. In Table \ref{tab:correctstemmedincorrect}, we can see consistent preferences for correct forms over stemmed forms on morphological relations. All examples shown are for GPT-J.
\begin{table}
\centering
\begin{tabular}{| l | l |}
\hline
\textbf{Subject} & \textbf{Jacobian Top-3} \\
\hline
society & societies, Soc, soc \\
child & children, children, Children \\
success & successes, success, Success \\
series & series, Series, Series \\
woman & women, women, Women \\
\hline
righteous & righteousness, righteous, \ldots \\
conscious & consciousness, conscious, \ldots \\
serious & seriousness, serious, serious \\
happy & happiness, happy, happy \\
mad & madness, mad, being \\
\hline
invest & investment, invest, investing \\
amuse & amusement, amuse, amusing \\
accomplish & accomplishment, accomplish, \ldots \\
displace & displacement, displ, dis \\
reimburse & reimbursement, reimburse, reimb \\
\hline
globalize & globalization, global, international \\
install & installation, install, Installation \\
continue & continuation, continu, contin \\
authorize & authorization, Authorization, \ldots \\
restore & restoration, restitution, re \\
\hline
manage & manager, managers, manager \\
teach & teacher, teachers, teach \\
compose & compos, composer, composing \\
borrow & borrower, lender, debtor \\
announce & announcer, announ, ann \\
\hline
\end{tabular}
\caption{\textbf{[noun\_plural], [verb+er], [verb+ment], [adj+ness], [verb+tion]} Selected examples of full subject tokens demonstrate that the linear Jacobian approximation captures irregular morphology effectively, reproducing both stemmed and full subject forms.}
\label{tab:full-forms}
\centering
\centering
\begin{tabular}{|l|c|}
\hline
\textbf{Relation} & \textbf{\# Unique} \\ \hline
\textbf{un+adj} & \textbf{7} \\ \hline
\textbf{over+adj} & \textbf{4} \\ \hline
\textbf{re+verb} & \textbf{15} \\ \hline
name - nationality & 13 \\ \hline
animal - shelter & 18 \\ \hline
synonyms - intensity & 35 \\ \hline
verb+able & 47 \\ \hline
noun - plural & 47 \\ \hline
\end{tabular}
\caption{The number of unique start tokens for correct objects across selected BATS relations. Start tokens which occur frequently among objects indicate a non-injective subject-object map, making linear approximation a less suitable choice as an approximator.}
\label{tab:unique start tokens}
\vspace{0.2in}
\begin{tabular}{|c|c|c|}
\hline
\textbf{Correct} & \textbf{Stemmed} & \textbf{Incorrect} \\ \hline
42    & 0       & 0        \\ \hline
23    & 11      & 9        \\ \hline
7      & 35      & 6        \\ \hline
\end{tabular}
\caption{Correct, stemmed, and incorrect suffix counts for \textbf{[noun\_plural]}, \textbf{[verb+tion]} and \textbf{[adj+ness]} from the top prediction of a fixed layer Jacobian approximation further suggests consistent linear encoding beyond stemmed forms.}
\label{tab:correctstemmedincorrect}
\hspace{5pt} 
\end{table}

\section{Bias Results demonstrate W necessity}

A comparison of linear and affine approximators against the bias approximator demonstrates that the bias term $b_r$ alone cannot explain the relational encoding but contributes alongside the Jacobian $W_r$. This suggests that these operations play complementary roles in semantic and encyclopedic relations.

\begin{figure*}[htbp]
    \centering
    \includegraphics[width=1\linewidth]{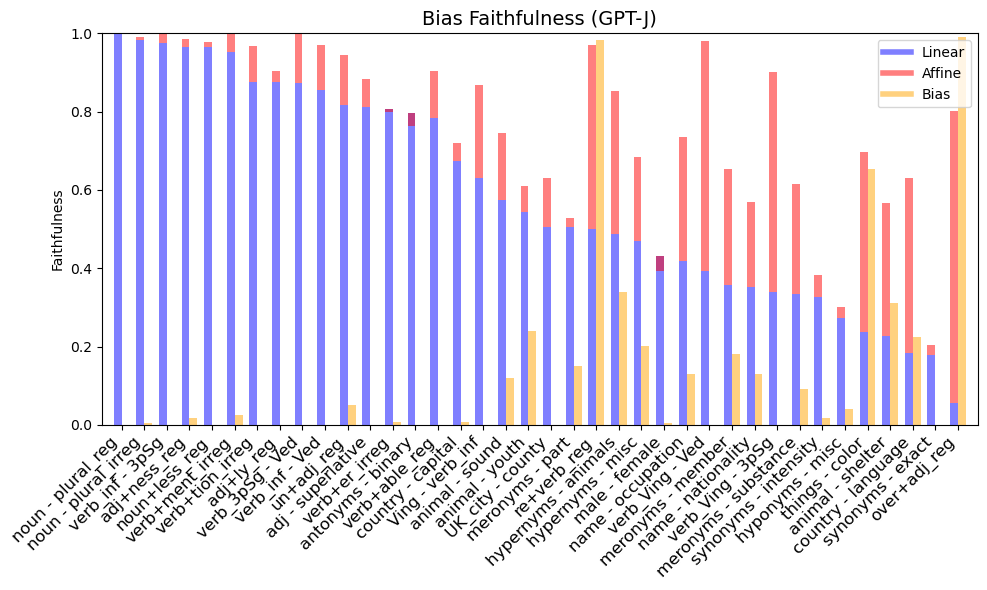}
    \caption{A comparison of the affine LRE against the Bias approximator demonstrates the necessity of the multiplicative (Jacobian) operator. Across semantic and encyclopedic relations, the additive Bias operator exhibits far better performance on morphology, providing evidence for complementary additive and multiplicative mechanisms.}
    \label{fig:bias}
    \includegraphics[width=1\linewidth]{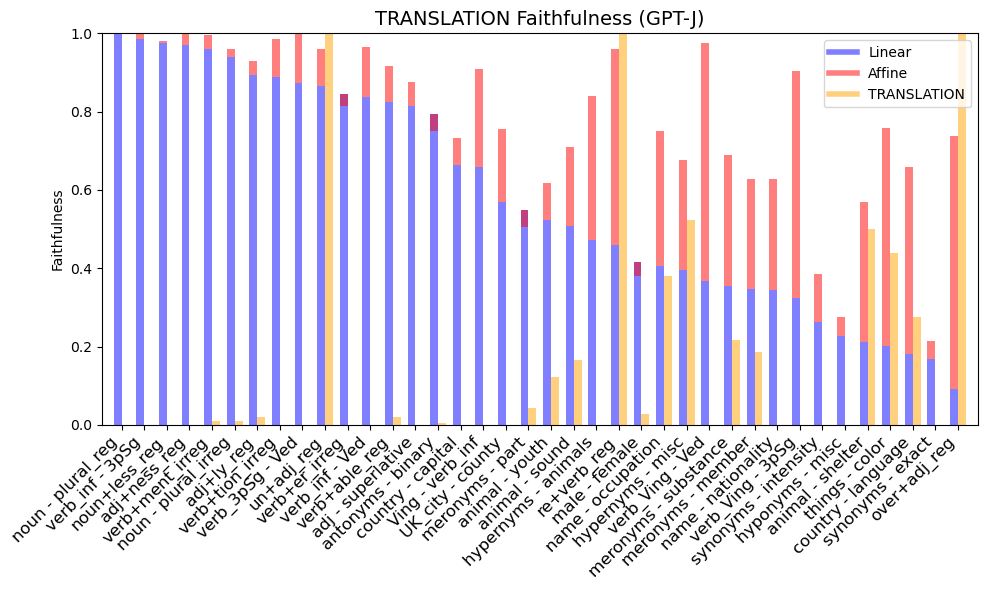}
        \caption{The TRANSLATION approximator $\tilde{\textrm{\textbf{o}}} = \text{Bias({\textbf{s})}} = \textbf{s} + b_r$, with $b_r = \mathbb{E}(\textbf{o} - \textbf{s})$, performs well on semantic and encyclopedic relations, similar to the Bias approximator.}
    \label{fig:translation}
\end{figure*}

The TRANSLATION operator, inspired by Merullo et al. \shortcite{Mer:23} and vector arithmetic, is also additive and performs similarly to the Bias operator. Figure \ref{fig:translation} demonstrates the additive TRANSLATION approximator against both the affine and linear LRE. Like the bias approximator, the TRANSLATION approximator succeeds when the gap between the Jacobian and LRE is large. This suggests that semantic information plays a crucial role in bridging some subject-object relations.

\section{Linear Projection}

\begin{figure*}[htbp!]
    \centering
    \includegraphics[width=1\linewidth]{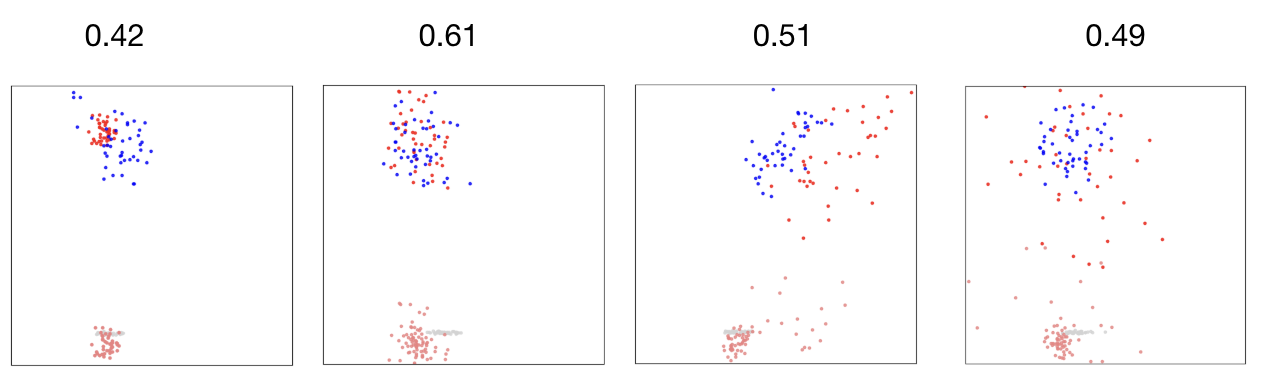}
    \caption{Projected subspace distances for fifty approximated object states \(\textcolor{red}{\beta W \mathbf{s} + b_r}\) and true object states \(\textcolor{blue}{\mathbf{o}}\) for $\textbf{[animal - youth]}$. The subspace used is $\{\bot, b_r\}$, where $\bot$ is a randomly chosen orthogonal vector to $b_r$. The faithfulness scores of each relation are displayed above. With $\beta$ values of 1, 3, 5, and 7, the hyperparameter $\beta$ is shown to be crucial for faithful approximation in the affine LRE.}
    \label{fig:proj}
\end{figure*}

We find that linear projection to $\mathbb{R}^2$ can yield interpretable geometric representations. Specifically, we use a basis of the bias vector $b$ and a random normalized vector, which has been orthogonalized with Gram-Schmidt to $b$, and compare approximated transformations against true object states. As seen in Figure \ref{fig:proj}, we find subspace distance corresponding heavily to faithfulness. Additionally, we validate that the $\beta$ hyperparameter is necessary for recovering scale lost in layer normalization, as conjectured by \cite{Her:23}.

We project approximations \textcolor{gray}{\textbf{s}}, \textcolor{magenta}{$\beta W \textbf{s}$}, \textcolor{red}{$\beta W \textbf{s} + b$ }, as well as a calculated hidden state for the correct object output \textcolor{blue}{$\textbf{o}$}.  These projections suggest $W$ is primarily responsible for transforming the underlying distribution to be geometrically similar to the output, while $b$ contributes the majority of movement in vector space. \\
The term $b_r$ could be compared to the vectors used by Mikolov and many others, and the concept vector subsequently formalized by Park. However, the bias vector and the concept vector are not truly analogous. The bias term describes an offset from the transformed subject to the object: $b_r = \mathbb{E}(o - W_r\textbf{s})$, not $b_r =\mathbb{E}(o - \textbf{s})$.  In practice, we find that bias and concept vectors are close in cosine similarity, and likely serve similar roles.

\begin{figure*}
\centering
\includegraphics[width=1\linewidth]{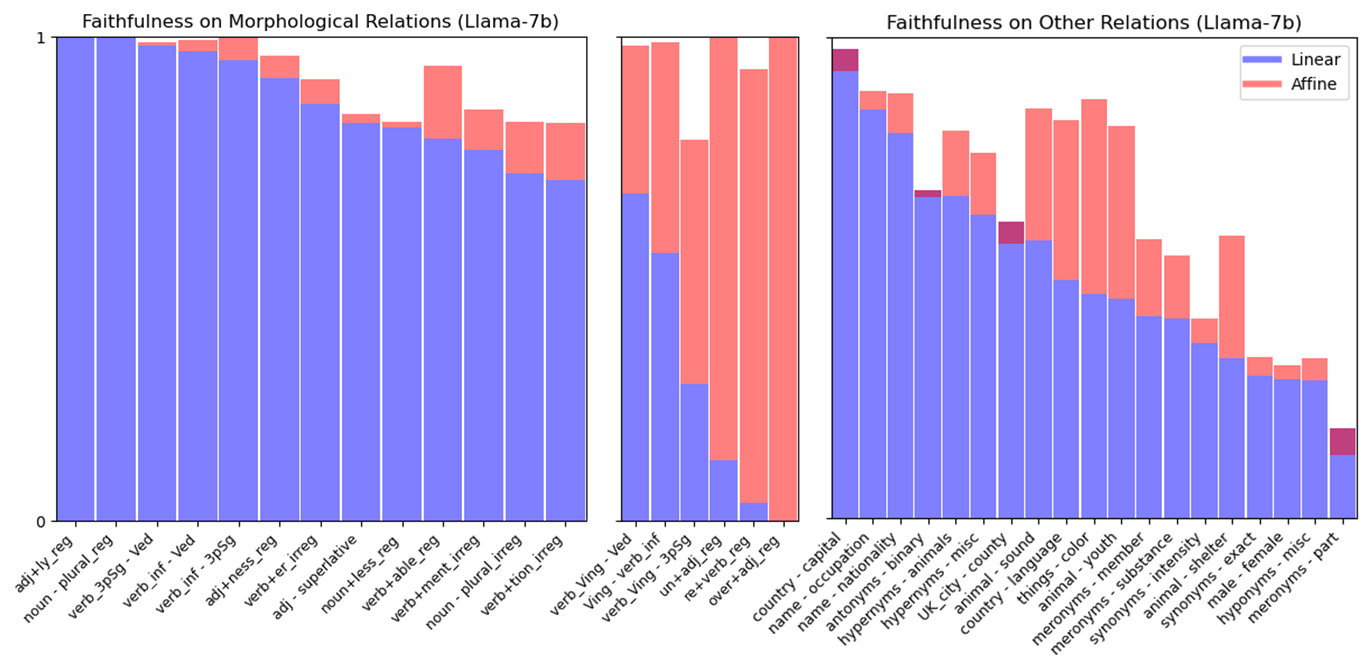}
\caption{Llama-7b results support a generalization across models: many morphological relations are linearly approximable, while semantic and encyclopedic relations benefit greatly from the affine method. Out of a range of subject layers 4-16, the best performing approximation is averaged ($n = 4$).}
\label{fig:faithfulness-llama}
\end{figure*}

\end{document}